\documentclass[letterpaper, 10 pt, conference]{ieeeconf} 

\IEEEoverridecommandlockouts      

\usepackage{amsmath}
\usepackage[utf8]{inputenc}
\usepackage{newunicodechar}
\usepackage{stfloats}
\usepackage{siunitx}
\usepackage{diagbox}
\usepackage{makecell}
\usepackage{cite}
\usepackage{ulem} 
\usepackage{balance}
\usepackage{graphicx}
\usepackage{float}
\usepackage{stfloats}
\usepackage{hyperref}
\usepackage{float}
\usepackage[ruled]{algorithm2e}
\usepackage{caption}
\usepackage{multirow}
\usepackage{booktabs}
\usepackage{indentfirst}
\usepackage{subcaption}
\usepackage{color}
\usepackage{lipsum}
\usepackage{amsmath,esint}
\usepackage{cuted}
\usepackage{CJKutf8}
\usepackage{verbatim}
\usepackage{bm}
\usepackage{setspace}
\usepackage{array,caption}
\usepackage{fancyhdr}
\usepackage{subcaption}

\hypersetup{
	colorlinks=true,
	linkcolor=blue,
	citecolor=blue,
	urlcolor=blue,
}

\title{\LARGE \bf
A Novel Robotic Variable Stiffness Mechanism Based on Helically Wound Structured Electrostatic Layer Jamming
}

\author{Congrui Bai, Zhenting Du, and Weibang Bai\textsuperscript{*} 
\thanks{
This work is supported by the Shanghai Pujiang Program under grant 23PJ1408500, by the Shanghai Frontiers Science Center of Human-centered Artificial Intelligence (ShangHAI), MoE Key Laboratory of Intelligent Perception and Human-Machine Collaboration (KLIP-HuMaCo). The experiments of this work were supported by the Core Facility Platform of Computer Science and Communication, SIST, ShanghaiTech University.
Corresponding author: Weibang Bai \textit{(wbbai@shanghaitech.edu.cn)}.}
\thanks{Congrui Bai and Weibang Bai are with the ShanghaiTech Automation and Robotics (STAR) Center, School of Information Science and Technology, ShanghaiTech University, Shanghai, 201210, China.}
\thanks{
Zhenting Du was with the ShanghaiTech Automation and Robotics (STAR) Center, School of Information Science and Technology, ShanghaiTech University, China. He is now with the King’s College London, London, U.K.}
}

\begin{document}

\maketitle
\thispagestyle{empty}
\pagestyle{empty}

\begin{abstract}

This paper introduces a novel variable stiffness mechanism termed Helically Wound Structured Electrostatic Layer Jamming (HWS-ELJ) and systematically investigates its potential applications in variable stiffness robotic finger design. The proposed method utilizes electrostatic attraction to enhance interlayer friction, thereby suppressing relative sliding and enabling tunable stiffness. Compared with conventional planar ELJ, the helical configuration of HWS-ELJ provides exponentially increasing stiffness adjustment with winding angle, achieving significantly greater stiffness enhancement for the same electrode contact area while reducing the required footprint under equivalent stiffness conditions. Considering the practical advantage of voltage-based control, a series of experimental tests under different initial force conditions were conducted to evaluate the stiffness modulation characteristics of HWS-ELJ. 
The results demonstrated its rational design and efficacy, with outcomes following the deduced theoretical trends. 

Furthermore, a robotic finger prototype integrating HWS-ELJ was developed, demonstrating voltage-driven stiffness modulation and confirming the feasibility of the proposed robotic variable stiffness mechanism.

\end{abstract}

\section{Introduction}

Recent advances in robotics have extended their applications to dynamic, uncertain scenarios that often involve complex human-robot interactions and demand greater adaptability and resilience. 
Consequently, robotic joints must not only provide high compliance to ensure safety and adaptability during physical interaction with humans or the environment, but also sufficient stiffness to support precise operations, heavy loads, and complex tasks. The balance between compliance and stiffness has thus become a central challenge in next-generation robotic joint design.

Conventional rigid joints exhibit high positional accuracy and strong load capacity, but their lack of inherent compliance poses potential safety risks during interactions with humans or the environment \cite{zhou2025unifiedmanipulabilitycomplianceanalysis}\cite{10606048}. In contrast, conventional flexible joints offer good resistance to impact and protection, but often compromise energy efficiency and dynamic responsiveness \cite{Zhu2024}. Therefore, achieving variable stiffness and dynamically switching between rigidity and compliance according to task requirements has emerged as a key issue in robotics research.

To address this challenge, a variety of variable stiffness technologies have been proposed, which can be broadly categorized into three groups. 
The first group consists of mechanically based approaches, such as series elastic actuators (SEA)\cite{ZHAO2024105541}, antagonistic actuators, and variable stiffness actuators (VSA)\cite{7330025}. These methods achieve modulation of stiffness through structural adjustments or spring preloading, offering a wide range of stiffness and straightforward modeling, but typically at the cost of increased complexity, size, weight, and energy consumption, which hinders miniaturization and integration. 
The second group involves smart materials, including Shape Memory Alloys (SMAs)\cite{HASSAN2024112861}, Electroactive Polymers (EAPs)\cite{polym17060746}, Magnetorheological or Electrochemical Fluids (MRF/ERF)\cite{LiquidRobotics}, and low-melting-point alloys\cite{LowMeltingAlloys}. These materials exhibit advantages in compactness and integration but are constrained by limitations in response speed, energy efficiency, thermal sensitivity, and durability. 
The third group encompasses friction or damping-based strategies\cite{act9040104}, such as particle jamming\cite{7814318}, fiber jamming\cite{10252027}, and layer jamming\cite{CARUSO2023108325}. Particle jamming achieves stiffness modulation by regulating external pressure and has been widely applied in soft joints and multi-finger grippers\cite{https://doi.org/10.1002/aisy.202400285}; however, it often requires large material volumes and offers limited spatial control of stiffness. Fiber jamming enhances stiffness by inter-fiber friction under compression and has shown promise in wearable haptic feedback systems, though its mechanical modeling remains underdeveloped\cite{doi:10.1089/soro.2019.0203}. In contrast, layer jamming offers advantages such as compactness, rapid response, and design flexibility, making it suitable for applications in soft fingers, robotic grippers\cite{Choi2019SoftMV}, and wearable devices \cite{act13020064}. 

Nonetheless, conventional layer jamming systems typically rely on vacuum-induced compression, which results in bulky structures with stringent sealing requirements, thereby hindering miniaturization and integration in compact systems.
Against this backdrop, Electrostatic Layer Jamming (ELJ) has been introduced\cite{8613892}. By applying a high-voltage electrostatic field between flexible electrode layers, ELJ generates attractive forces that enhance interfacial friction, enabling continuously tunable stiffness. Compared with pneumatic jamming, ELJ offers fast response, lightweight design, thin profile, and ease of integration, rendering it particularly suitable for miniaturized and wearable robotic systems. However, most current ELJ mechanisms are predominantly designed in planar stacked configurations, which typically require either large electrode areas or multilayer assemblies to achieve sufficient stiffness. This design results in poor space utilization and limits their applicability in compact joint structures.

To address this limitation, this study proposes a Helically Wound Structured ELJ (HWS-ELJ), where flexible electrodes are helically wound around a cylindrical core, thereby significantly improving space utilization while leveraging the mechanical characteristics of the helical configuration to enhance stiffness modulation. Based on the Euler belt friction model [20], increasing the winding angle leads to an exponential increase in interfacial friction, thereby greatly augmenting the stiffness adjustment capability. Moreover, the helically wound form improves geometric adaptability, facilitating integration with joints and bending structures, while preserving the inherent advantages of ELJ technology, including rapid responsiveness, structural simplicity, and high reversibility, making it even suitable for application in flexible robotic fingers.
\begin{figure}[b]
    \centering
    \includegraphics[width=0.49\textwidth]{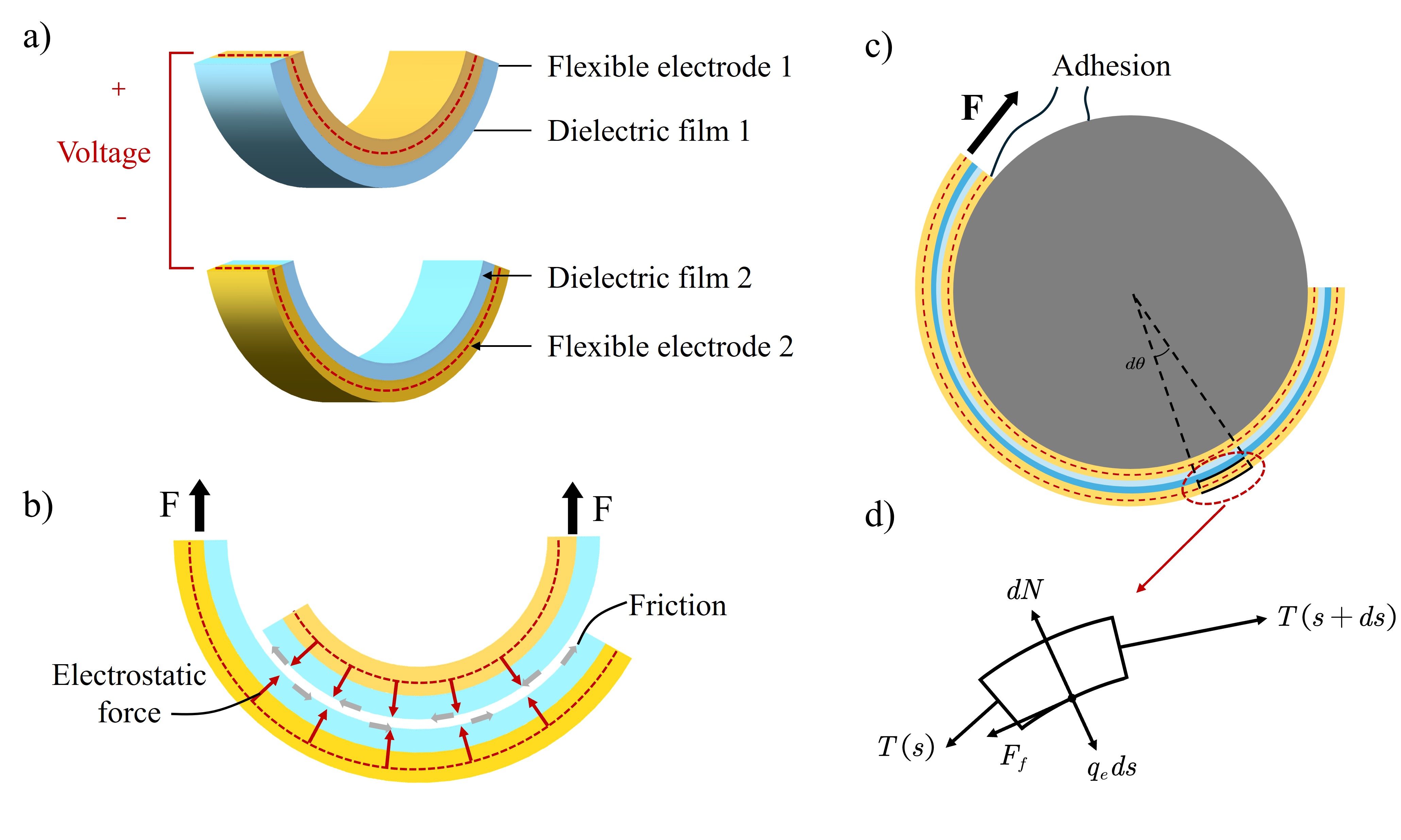}
    \vspace{-2mm}
    \caption{Structural schematic diagram of HWS-ELJ. a) Main structure: two electrode sheets with PI films attached. b) Principle of variable stiffness of the HWS-ELJ. c) Force on a unit electrode during relative sliding. d) Diagram of force analysis on an infinitesimal section of the wound electrode.}
    \label{fig:structure_diagram}
\end{figure}

\section{Variable Stiffness Mechanism Analysis}
\subsection{Structure of HWS-ELJ} 
The helically wound structured electrostatic layer jamming (HWS-ELJ) represents a novel configuration derived from the principle of electrostatic layer jamming (ELJ). The underlying mechanism of ELJ is that when two parallel electrodes are energized, the resulting electrostatic attraction generates additional interfacial friction, thereby effectively suppressing relative motion and enhancing the overall stiffness of the system. Building upon this principle, this study proposes a helically wound structure, as illustrated in Fig.~\ref{fig:structure_diagram}(a) and Fig.~\ref{fig:structure_diagram}(b). The HWS-ELJ consists of two flexible electrodes helically wound around the surface of an insulating cylindrical core, with each electrode composed of a dielectric thin film and a metallic flexible sheet. One electrode is firmly attached to the cylinder surface, while the other is wound helically to maintain continuous contact. Under this configuration, when an initial tensile force is applied to the winding electrode and it continues to wrap around the core, the terminal force of the electrode increases exponentially with the winding angle. The two electrodes are connected to the positive and negative terminals of a high-voltage power supply, respectively. Upon the application of high voltage, the electrostatic attraction between the electrodes not only increases the normal pressure but also significantly enhances the interfacial friction. Since the electrostatic force is directly dependent on the applied voltage, the interlayer friction and stiffness of the HWS-ELJ system can be actively and continuously modulated.

\begin{figure}[h]
    \centering
    \includegraphics[width=0.8\linewidth]{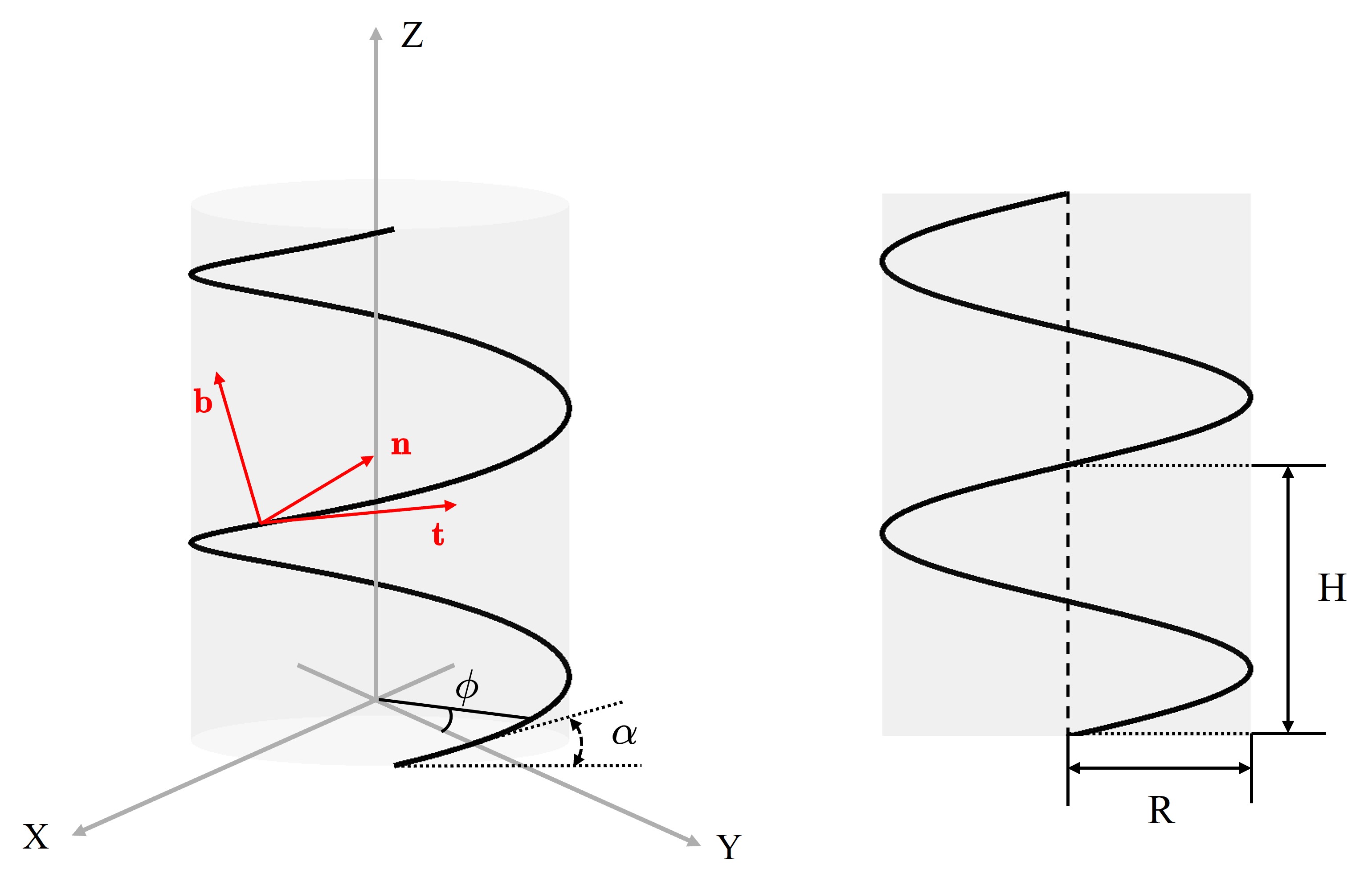}
    \caption{Schematic of the Geometric Parameters of the HWS-ELJ}
    \vspace{-3mm}
    \label{fig:Spiral line}
\end{figure}
\subsection{Modeling the HWS-ELJ}
\label{sec:model}
We focus on the HWS-ELJ, conducting theoretical analysis to investigate the relationship between applied voltage and the interfacial friction force, as shown in Fig.~\ref{fig:structure_diagram}(c). The structure consists of a flexible electrode strip with embedded conductive layers, having a width denoted as $\omega$, helically wound around the surface of a cylindrical substrate with radius R, the spiral structure is shown in Fig.~\ref{fig:Spiral line}. The outer surface of the cylinder is partially covered with corresponding conductive electrodes only in regions that overlap with the wound electrode strip, while the remaining surface is kept electrically insulated. To ensure electrical isolation and effective electrostatic adhesion, the opposing electrode surfaces are both coated with a dielectric film, forming a stable capacitive structure. During the winding process, the electrodes are arranged along the axial direction with a constant helical pitch $H$, satisfying H$\textgreater$$\omega$. This condition ensures that no interlayer overlap occurs on the cylindrical surface, resulting in a uniformly spaced helical structure. Let the total winding angle be denoted as $\Phi$.
Establish the spiral line parametric equation:
\begin{equation}    
\mathbf{r}\left( \phi \right) =\left( R\cos \phi ,R\sin \phi ,\frac{H}{2\pi}\phi \right) ,\ \ \ \phi \in \left[ 0,\varPhi \right] 
\end{equation}

The curvature $\kappa$ of and torsion $\tau $ of the helix are \cite{Konyukhov2013}:

\begin{equation}
\kappa = \frac{R}{R^2 + \left(\frac{H}{2\pi}\right)^2}
\end{equation}
\begin{equation}
\tau = \frac{H/2\pi}{R^2 + \left(\frac{H}{2\pi}\right)^2}
\end{equation}

Conduct a mathematical analysis on an infinitesimal segment extracted from the helical electrode. The differential arc length ds is expressed as:
\begin{equation} 
ds=\sqrt{dx^2+dy^2+dz^2}=\sqrt{R^2+\left( \frac{H}{2\pi} \right) ^2}d\phi =ad\phi 
\end{equation}
\begin{equation} 
a=\sqrt{R^2+\left( \frac{H}{2\pi} \right) ^2}
\end{equation}

To characterize the geometric and mechanical properties along the curve, the natural Serret--Frenet coordinate system \cite{1944A}is introduced. This local frame consists of the tangent vector $\mathbf{t}$, the principal normal vector $\mathbf{n}$, and the binormal vector $\mathbf{b}$. The tangent vector $\mathbf{t}$ points along the curve in the forward direction, the normal vector $\mathbf{n}$ points toward the center of curvature, and the binormal vector $\mathbf{b}$, determined by the right-hand rule, is perpendicular to the osculating plane. Together, they form an orthonormal basis that facilitates the description of local geometry and mechanical behavior. The basis vectors are as follows:

\begin{equation} 
\mathbf{t}=\frac{d\mathbf{r}}{ds}=\frac{1}{a}\left( -R\sin \phi ,R\cos \phi ,\frac{H}{2\pi} \right) 
\end{equation}
\begin{equation} 
\mathbf{n}=\frac{1}{\kappa}\frac{d\mathbf{t}}{ds}=\left( -\cos \phi ,-\sin \phi ,0 \right) 
\end{equation}
\begin{equation} 
\mathbf{b}=\mathbf{t}\times \mathbf{n}=\frac{1}{a}\left( \frac{H}{2\pi}\sin \phi ,-\frac{H}{2\pi}\cos \phi ,R \right) 
\end{equation}

To accurately characterize the mechanical behavior during the electrode winding process, we analyze a differential electrode segment of length $ds$ adhered to the surface of the cylindrical substrate. As shown in Fig.~\ref{fig:structure_diagram}(d), within this infinitesimal region, the electrode is subjected to four main types of forces: internal tension, normal support from the substrate, friction, and electrostatic adhesion force. Considering that the electrode moves along a helical path at a constant velocity, the system can be regarded as being in a quasi-static equilibrium in both the tangential and normal directions.

The internal tension is distributed along the tangential direction of the electrode. At the left end of the segment (position $s$), the tension is expressed as:
\begin{equation}
\mathbf{T}(s) = T(s) \, \mathbf{t}
\end{equation}
where $\mathbf{t}$ is the unit vector in the tangential direction. At the right end (position $s + ds$), the tension can be expanded using a Taylor series as:
\begin{equation}
\mathbf{T}(s + ds) \approx \mathbf{T}(s) + \frac{d\mathbf{T}}{ds} ds
\end{equation}
The net force exerted on the segment by tension can thus be expressed as:
\begin{equation}
\mathbf{T}(s + ds) - \mathbf{T}(s) \approx \frac{d\mathbf{T}}{ds} ds
\end{equation}
which acts in the direction opposite to the rate of change of tension.

The substrate provides normal support force for the electrode. Assuming that the direction of this force is opposite to $\mathbf{n}$, the model of this force is:
\begin{equation}
\mathbf{F}_{\mathbf{N}}=-dN\mathbf{n}
\end{equation}
where $\mathbf{n}$ is the unit normal vector and $dN$ is the differential normal force magnitude.

The friction force opposite to the motion of the differential electrode segment is proportional to the normal force, which can be expressed as follows:
\begin{equation}
\mathbf{F}_{\mathbf{f}}=-\mu dN\mathbf{t}
\end{equation}
with $\mu$ being the coefficient of friction.

When a voltage is applied, an electrostatic attraction will be generated between the two electrodes. Assuming that this electrostatic attraction is uniformly distributed across the contact area, the system can be analyzed using the parallel-plate capacitor model. Considering an electrode of width $\omega$, the combined effect of electrostatic attraction and the initial preload ensures that the two electrodes remain in close contact. Consequently, any potential air gap between the electrodes can be neglected in the theoretical analysis, and the effective inter-electrode distance $d_e$ is defined as the sum of the dielectric layer thicknesses attached to their surfaces. 

\begin{figure*}[t]
    \centering
    \includegraphics[width=0.98\textwidth]{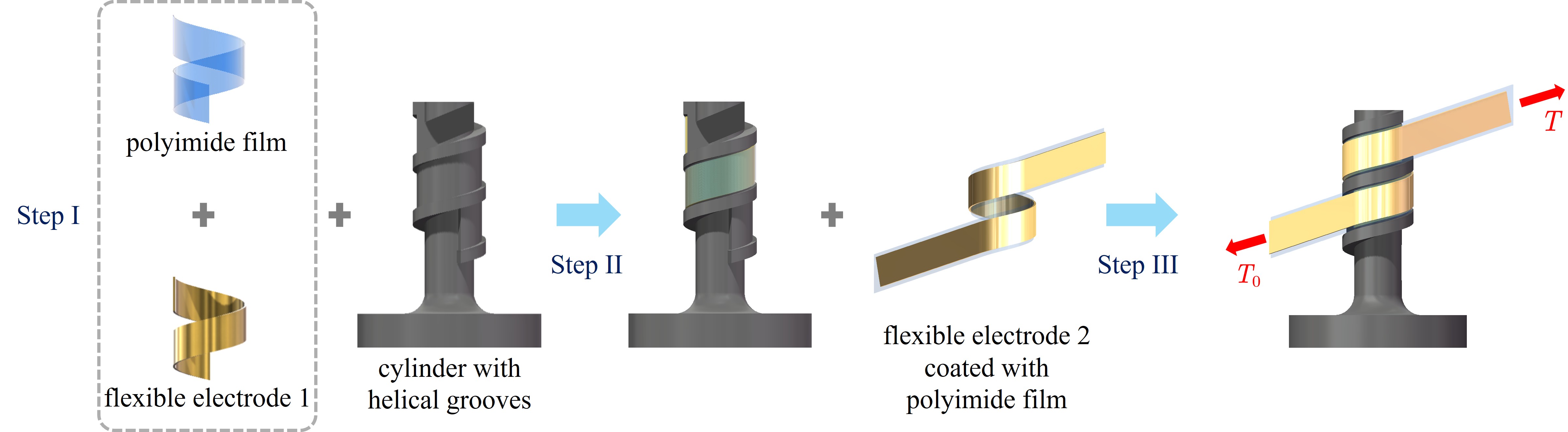}
    \caption{Fabrication process of the prototype. Step I: Attach the dielectric film to the flexible electrode 1. Step II: Attach the flexible electrode 1 coated with dielectric film to the cylinder with helical grooves. Step III: Wrap the flexible electrode 2 coated with dielectric film around the cylinder with helical grooves.}
    \label{fig:Fabrication process}
\end{figure*}

Specifically, Electrode~1 is coated with a dielectric film characterized by permittivity $\varepsilon_1$ and thickness $d_1$, while Electrode~2 is covered with a dielectric film of permittivity $\varepsilon_2$ and thickness $d_2$. 
Since the thickness of the copper foil electrodes is negligible compared with that of the dielectric films, it is excluded from the model. 
Therefore, the total effective distance $d_e$ between the electrode cores can be expressed as:
\begin{equation} 
d_e = d_1 + d_2
\end{equation}

The electrostatic force per unit length is expressed as \cite{1998Classical}:
\begin{equation}
q_e = \frac{\varepsilon_{\text{0}}\varepsilon_{\text{e}}\omega V^2}{2 d_{\text{e}}^2}
\end{equation}
and thus the total electrostatic force on the segment is:
\begin{equation}
\mathbf{F}_{\mathbf{e}}=q_e ds\mathbf{n}
\end{equation}

The equivalent permittivity between the two electrodes is:
\begin{equation} 
\varepsilon _e=\frac{\varepsilon _1\varepsilon _2d}{\varepsilon _1d_2+\varepsilon _2d_1}
\end{equation}

According to Newton's second law, under the condition of uniform motion, the system exhibits zero acceleration, implying that the infinitesimal segment is in a state of static equilibrium. Consequently, the sum of all external forces acting on the element must be 0:
\begin{equation} 
\frac{d\mathbf{T}}{ds}ds+\left( -dN+q_eds \right) \mathbf{n}-\mu dN\mathbf{t}=0
\end{equation}

Divide both sides by $ds$ simultaneously and expand the tension derivative:
\begin{equation} 
\left( \frac{dT}{ds}\mathbf{t}+T\frac{d\mathbf{t}}{ds} \right) +\left( -\frac{dN}{ds}+q_e \right) \mathbf{n}-\mu \frac{dN}{ds}\mathbf{t}=0
\end{equation}
According to the Serret-Frenet formula\cite{1944A}, $\frac{dt}{ds}=\kappa \mathbf{n}$, where $\kappa$ is the curvature. Thus:
\begin{equation} 
\left( \frac{dT}{ds}\mathbf{t}+\kappa T\mathbf{n} \right) +\left( -\frac{dN}{ds}+q_e \right) \mathbf{n}-\mu \frac{dN}{ds}\mathbf{t}=0
\end{equation}

The force in the above equation can be decomposed into tangential and normal components. Accordingly, the equation can be expanded along these two directions as follows:\\
Tangential component:
\begin{equation} 
\frac{dT}{ds}-\mu \frac{dN}{ds}=0\ \ \ \Rightarrow \ \ \ \frac{dT}{ds}=\mu \frac{dN}{ds}
\end{equation}
Normal component:
\begin{equation} 
\kappa T-\frac{dN}{ds}+q_e=0\ \ \ \Rightarrow \ \ \ \frac{dN}{ds}=\kappa T+q_e
\end{equation}
Accordingly, we obtain:
\begin{equation} 
\frac{dT}{ds}-\mu \kappa T=\mu q_e
\end{equation}
To solve this first-order linear ordinary differential equation, we construct the integration factors $I\left( s \right) =e^{-\mu \kappa s}$. Upon solving, we obtain:
\begin{equation} 
T=T_0e^{\mu \kappa s}+\frac{\varepsilon _0\varepsilon _e\omega V^2}{2d_{e}^{2}\kappa}\left( e^{\mu \kappa s}-1 \right) 
\end{equation}

\vspace{2mm}
\section{Variable Stiffness Mechanism Verification}
According to the theoretical model established in \ref{sec:model} section, the magnitude of the terminal tensile force $T$ is influenced by multiple factors, including the dielectric constant of the material, electrode area, initial pulling force, applied voltage, dielectric film thickness, and the curvature of the spiral winding of the electrode. Among these factors, voltage control is considered the most efficient and practical approach, as it allows convenient and continuous modulation of stiffness. Based on this consideration, the present experiment focuses on verifying the effect of voltage on the terminal tensile force. Furthermore, to ensure the generality of the experimental results, different tensile loads are applied at the end of the wound electrode, and the relationship between the terminal tensile force and the applied voltage is systematically investigated.

\subsection{Design and Manufacture of Prototypes} 
The design and fabrication process of the variable tensile stiffness (HWS-ELJ) specimens is illustrated in Fig.\ref{fig:Fabrication process}. The structure mainly consists of three components: copper foil, polyimide (PI) films, and a 3D-printed cylinder with helical grooves. Cylindrical specimens with a fixed helically wound angle of 450° were fabricated using 3D printing technology. The helical grooves were designed with a constant angle to ensure that the electrode maintained a contact angle of 450° during winding, providing precise guidance for the flexible electrode placement. For the fabrication of the HWS-ELJ specimens, an 8 mm-wide copper foil was first adhered along the helical groove of the cylindrical surface as the fixed electrode. A 10 mm-wide PI film was then laminated over the copper, extending 1 mm beyond each side to provide insulation and edge protection. The winding electrode was prepared in a similar manner: a 7 mm-wide copper foil was centrally placed on a 9 mm-wide PI film, with an identical PI layer laminated on the back to form a sandwich structure. This configuration enhanced the mechanical strength and durability of the electrode during the winding process.

To characterize the frictional properties, the coefficient of friction between two PI films was measured. Specifically, one PI film was smoothly attached to the surface of a table, and another PI film of varying weights (50 g, 100 g, and 200 g) was placed on top. A horizontal pulling force was applied to the upper film using a spring dynamometer. The force recorded at the onset of sliding was taken as the friction force, and the coefficient of friction was calculated by dividing this value by the corresponding weight. Each test was repeated three times to ensure reliability, and the specimen parameters are summarized in Table.~\ref{tab:specimen_parameters}.

\begin{table}[h]
    \centering
    \caption{Parameters of the Test Specimens}
    \label{tab:specimen_parameters}
    \resizebox{0.99\columnwidth}{!}{%
    \begin{tabular}{ccccc}
    \hline\hline
    \makecell{Specimen\\number} & 
    \makecell{Permittivity\\(F/m)} & 
    \makecell{Effective\\width\\(mm)} & 
    \makecell{Initial\\force\\(N)} & 
    \makecell{Coefficient\\of static\\friction} \\
    \hline
    1 & $3.6\times 8.85\times10^{-12}$ & 7.0 & 0.25 & 0.22 \\
    2 & $3.6\times 8.85\times10^{-12}$ & 7.0 & 0.50 & 0.22 \\
    3 & $3.6\times 8.85\times10^{-12}$ & 7.0 & 1.00 & 0.22 \\
    \hline\hline
    \end{tabular}%
    }
\end{table}

\begin{figure}[h]
    \centering
    \includegraphics[width=0.5\textwidth]{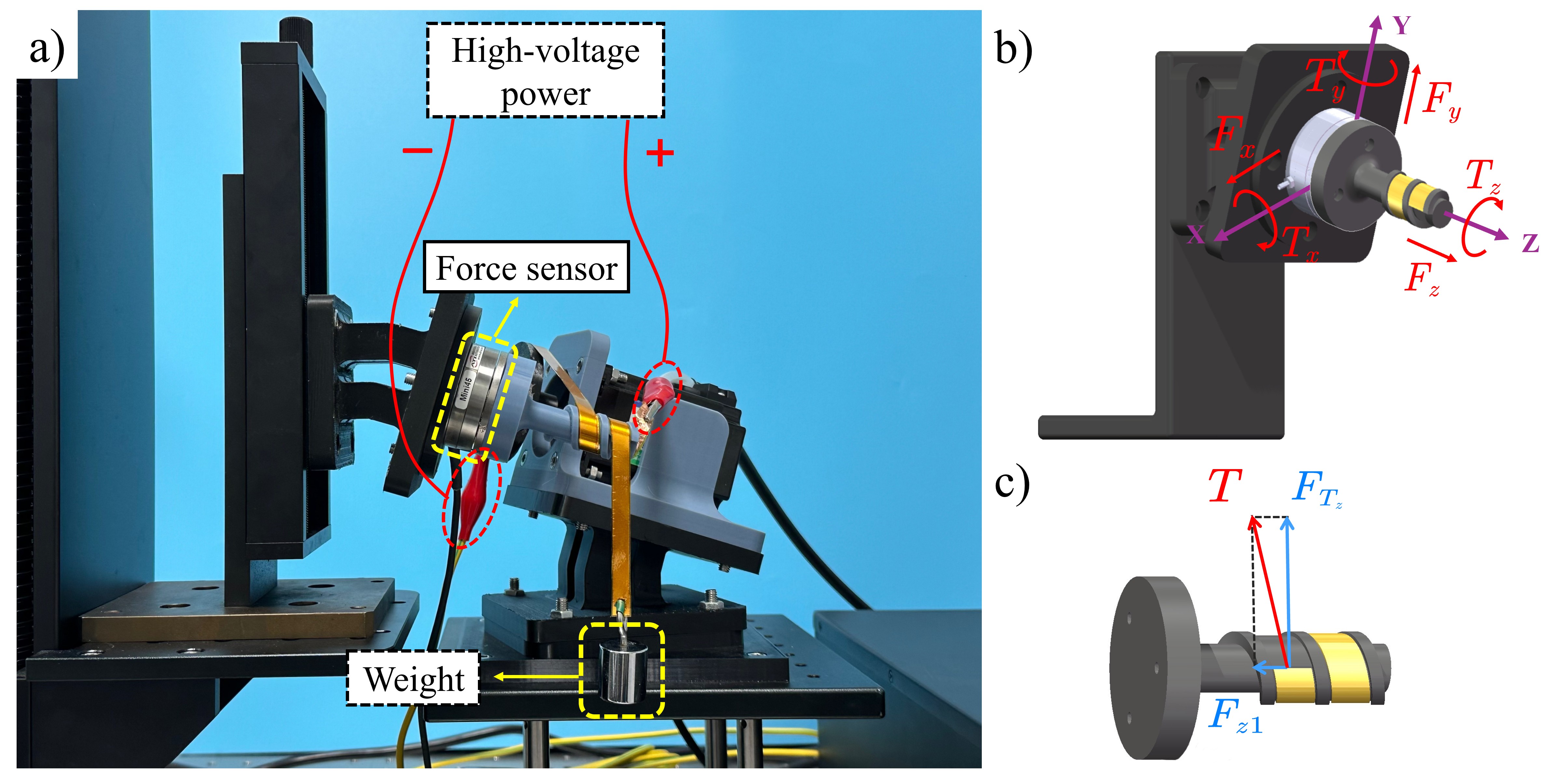}
    \caption{A tensile test of uniform relative sliding under a series of voltages. a) Validation experiment setup. b) Schematic diagram of force acquisition with a 6-axis force sensor. c) 
    Relationship between the force sensor output and the end tension.}
    \vspace{-2mm}
    \label{fig:setup}
\end{figure}

In this experiment, to investigate the influence of applied voltage on the terminal tensile force, weights of 25g, 50g, and 100g were respectively attached to the free end of the electrode under identical experimental conditions, and the corresponding terminal tensile forces at the wound electrode end were measured at different voltages. The validation experiment setup is shown in Fig.~\ref{fig:setup}(a), the cylindrical specimen with the fixed electrode was mounted on an ATI six-axis force sensor, which measured forces and torques in the x, y and z directions in real time, as shown in Fig.~\ref{fig:setup}(b). The sensor was secured to a lifting platform, allowing for adjustments in height. To ensure that the free end of the winding electrode remained vertical, the cylindrical specimen was oriented such that its terminal groove pointed downward, thereby guaranteeing that the initial tensile load was solely determined by the suspended weight. Prior to testing, the force sensor was calibrated to eliminate zero-offset errors and ensure measurement accuracy.

The orientation and height of the electrode were then adjusted so that the winding electrode was tangential to the motor flange after winding and fixed in place. During the experiment, the motor rotated at a constant speed, while the fixed and winding electrodes were connected to the positive and negative terminals of a high-voltage power supply, respectively. Once the motor was activated, real-time tensile force data were continuously recorded by the sensor to analyze the effective stiffness response of the HWS-ELJ specimens under different initial load conditions.

\begin{figure}[h]
    \centering
    \begin{subfigure}{0.8\linewidth}
        \centering
        \includegraphics[width=\linewidth]{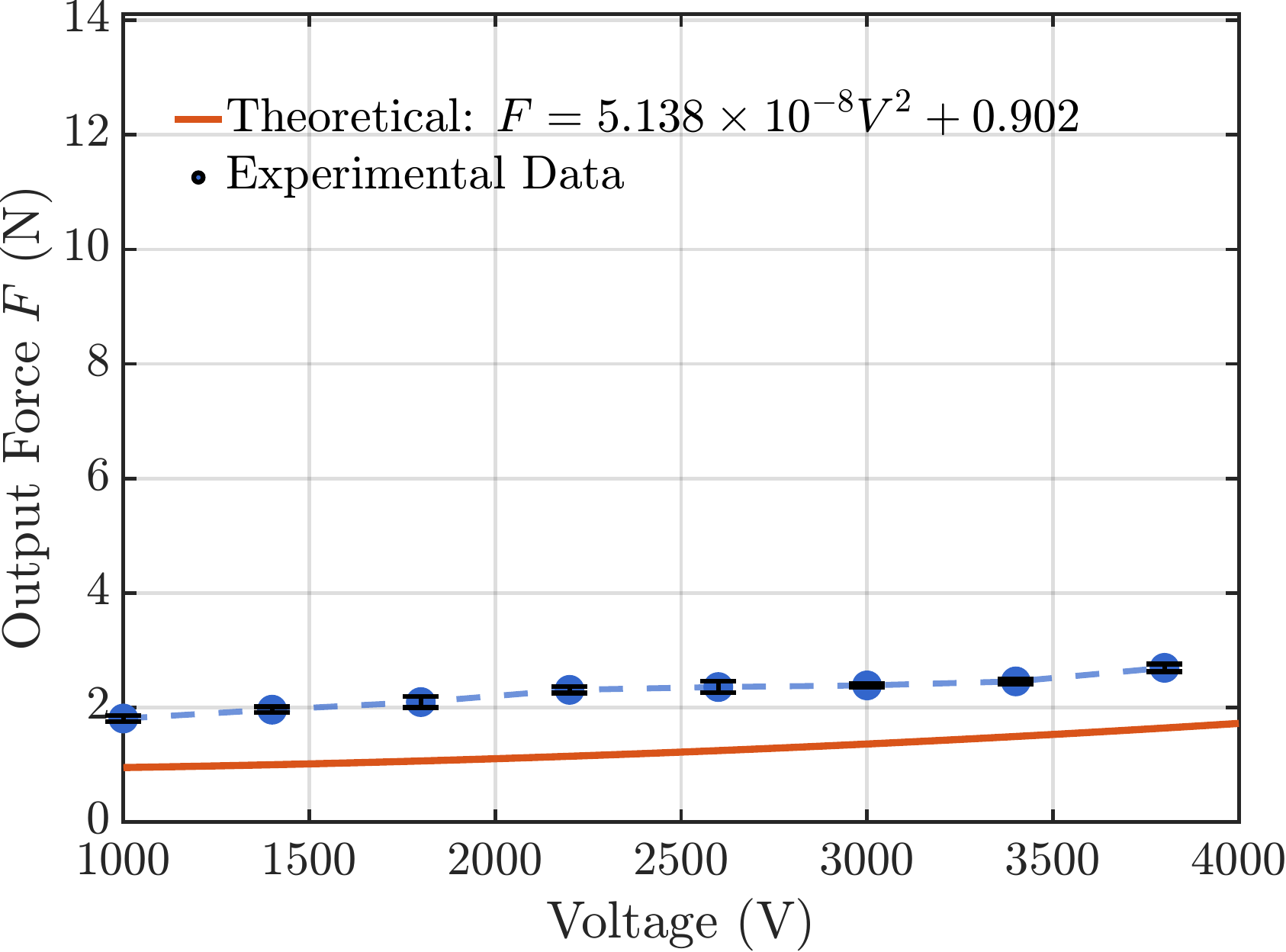}
        \vspace{-5mm}
        \caption{Load with a 25g weight}
        \vspace{2mm}
        \label{fig:AngleVoltage1}
    \end{subfigure}
    \begin{subfigure}{0.8\linewidth}
        \centering
        \includegraphics[width=\linewidth]{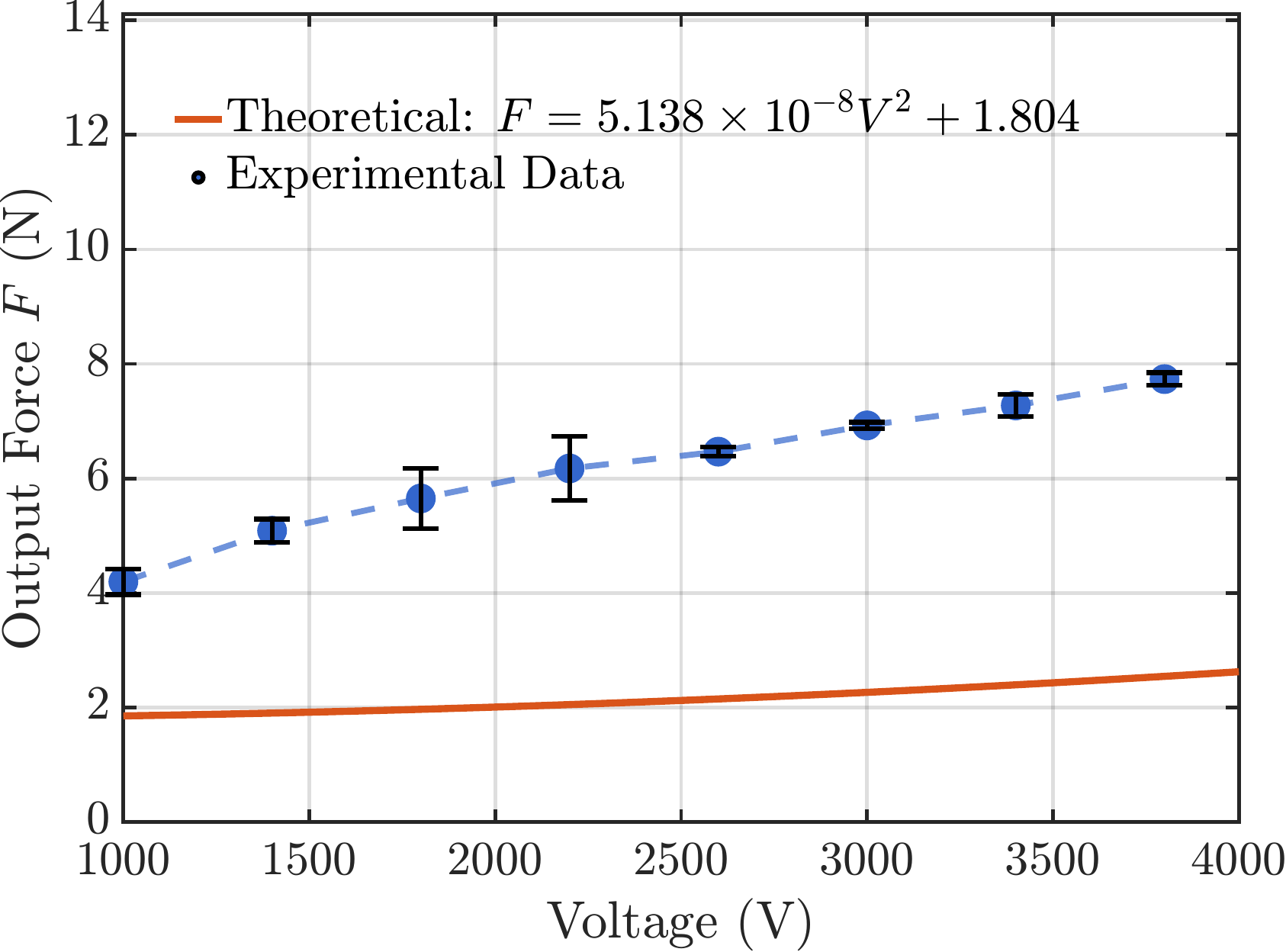}
        \vspace{-5mm}
        \caption{Load with a 50g weight}
        \vspace{2mm}
        \label{fig:AngleVoltage2}
    \end{subfigure}
    \begin{subfigure} {0.8\linewidth}
        \centering
        \includegraphics[width=\linewidth]{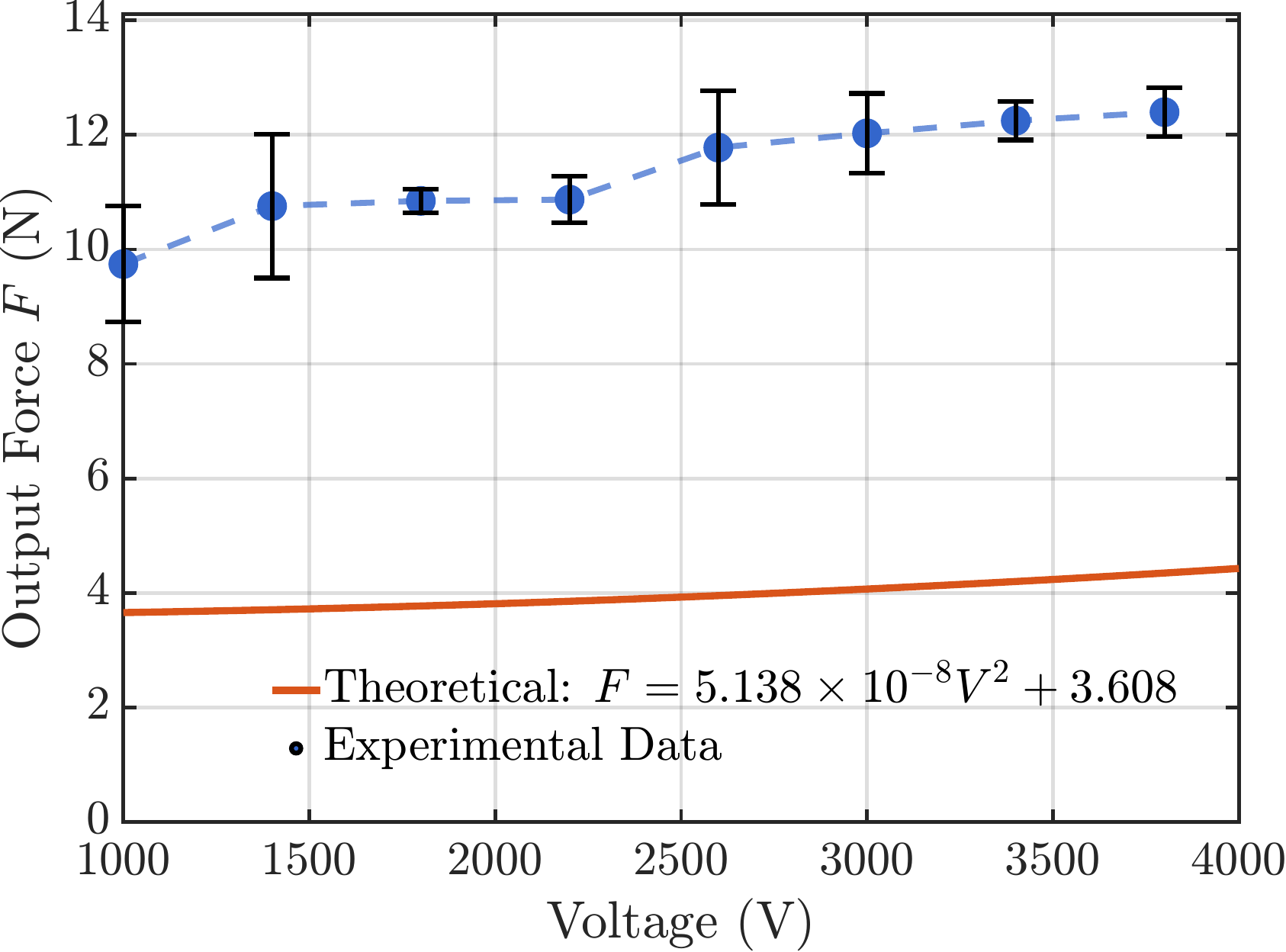}
        \vspace{-5mm}
        \caption{Load with a 100g weight}
        \vspace{2mm}
        \label{fig:AngleVoltage3}
    \end{subfigure}
    \caption{Comparison of Output Force vs. Voltage Under Various Loads for the HWS-ELJ.}
    \label{fig:DATA}
    \vspace{-4mm}
\end{figure}

\begin{figure*}[h]
    \centering
    \includegraphics[width=0.98\textwidth]{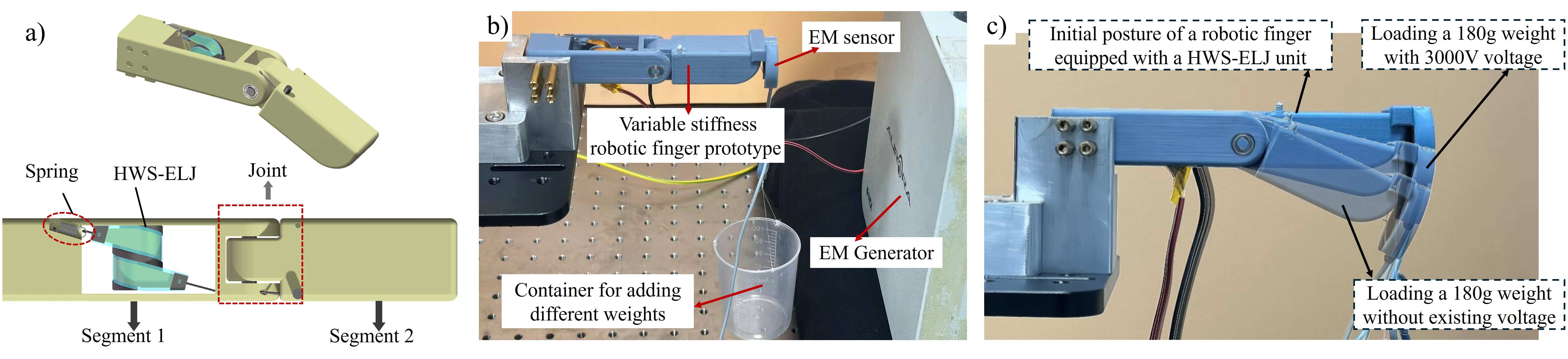}
    \caption{Application of the proposed HWS-ELJ to design a variable stiffness robotic finger. a) The 3D Model of the designed variable stiffness robotic finger. b) Measurement setup of the finger's stiffness with different voltages and different weights. c) Example of variable stiffness tests.}
    \label{fig:application}
\end{figure*}

\subsection{Data Processing}
During the experiment, a motor was used to drive the flexible electrode to wind uniformly along the surface of a cylindrical substrate at a constant speed. Throughout this motion, a force sensor recorded the real-time contact force exerted on the cylinder at a sampling frequency of 10Hz. After data collection, the recorded dataset was processed to remove outliers caused by transient disturbances or system noise. The remaining valid data points were then averaged to obtain a representative force value under steady-state conditions.

The voltage applied to the electrode was gradually increased from 1000V to 3800V in increments of 400V. The configuration of the force sensor is shown in the accompanying figure; it is capable of measuring forces ($F_x$, $F_y$, $F_z$) and torques ($T_x$, $T_y$, $T_y$) along the x, y, and z axes, respectively. Fig.~\ref{fig:setup}(c) shows the schematic diagram of the end tension of the wound electrode, it can be decomposed into two components: $F_{z1}$ and $F_{z2}$, $F_{z1}$ is directly measured by the force sensor along the z-axis, on the other hand, since the end of the wound electrode is tangent to the cylinder groove, $F_{z2}$ can be calculated as the torque about the z-axis divided by the radius $r$ of the groove on the cylindrical surface. Therefore, the total magnitude of the friction is obtained as the sum of these two components:
\begin{equation} 
F_{z2}=\dfrac{T_z}{r}
\end{equation}
\begin{equation} 
F_f=\sqrt{F_{z1}^{2}+F_{z2}^{2}}
\end{equation}

The specific numerical values and parameters used in the experiment are substituted into the theoretically derived equations. The corresponding results are presented in the subsequent \ref{sec.AnaResults} section.

\subsection{Results Analysis}\label{sec.AnaResults}
In the HWS-ELJ specimen, individual material layers do not undergo elongation under tensile loading; instead, they slide relative to each other, while the interfacial friction resists this relative motion. Since stiffness is defined as a material's resistance to deformation under external force, the magnitude of the interfacial friction can be used as an effective measure of the stiffness of the HWS-ELJ specimen. The conversion of the test data into the calculation of the friction force has been provided in the previous section.

Fig.~\ref{fig:DATA} presents a comparison between the experimental measurements and the theoretical prediction of output force as a function of applied voltage. The experiment was repeated 5 times for each sample under the same conditions, and the mean and standard deviation were taken.. The experimental data points are shown as blue filled circles, while the red solid line represents the theoretical model described by the quadratic equation:
The initial force can be calculated by:
\begin{equation} 
T_0=mg
\end{equation}

When the attached weight has a mass of 25 grams:
\begin{equation*} 
T\left( V \right) =5.138\times 10^{-8}V^2+0.902
\end{equation*}

When the attached weight has a mass of 25 grams:
\begin{equation*} 
T\left( V \right) =5.138\times 10^{-8}V^2+1.804
\end{equation*}

When the attached weight has a mass of 25 grams:
\begin{equation*} 
T\left( V \right) =5.138\times 10^{-8}V^2+3.608
\end{equation*}

As shown in Fig.~\ref{fig:DATA}, the experimental force generally increases with voltage, displaying a trend that qualitatively agrees with the theoretical prediction. However, compared with the theoretical predictions, the experimental data exhibit a significant increase. This discrepancy can be attributed not only to the omission of electrode edge effects in the theoretical model but also to the substantial influence of atmospheric pressure. Specifically, each blocking layer remains flat, and good contact is established between adjacent surfaces. When an external voltage induces electrostatic attraction between the two surfaces, localized low-pressure regions are formed at the microscopic level of the contact interface. These localized pressure drops enhance the normal force during sliding, thereby leading to an increased friction \cite{8613892}.

The experimental results demonstrate that the stiffness of the HWS-ELJ structure can be effectively regulated through the application of external voltage, thereby confirming its variable-stiffness capability. Under the tested conditions, when the initial tensile force was set to 1 N, the terminal tensile force increased to over 12 N with voltage application, representing a twelvefold amplification. Notably, the specimen used in this study had an effective area of only $
7mm\times 16\pi$ mm and a wound volume of approximately$
16\pi mm^2\times 16.25mm$, yet still achieved significant stiffness enhancement. This finding highlights that the mechanism not only enables tunable stiffness but also exhibits remarkable potential for miniaturization, making it particularly suitable for robotic joints and other applications where space is constrained.

\section{Implementation and Application}
To verify the feasibility and effectiveness of the proposed HWS-ELJ mechanism in practical robotic joints, the mechanism was integrated into a single-finger prototype, and corresponding experiments were carried out. 

\subsection{Finger Prototype Design and Experimental Setup} 
The prototype of the finger is shown in Fig.~\ref{fig:application}(a), the robotic finger used in this study consists of two segments, denoted Segment 1 and Segment 2. Segment 1 is fixed to the experimental platform, and the two segments are connected via a shaft and bearing. The entire structure is fabricated using 3D printing technology. Segment 1 is designed as a hollow structure, allowing the integration of the HWS-ELJ mechanism inside. A groove is machined near the fixed end of Segment 1 to accommodate a spring. Since the spring diameter is slightly smaller than that of the groove, its elongation and compression are not affected by friction constraints. An end of the spring is anchored to segment 1, providing the HWS-ELJ mechanism with an initial preload. In the absence of the wound electrode, segment 2 remains free to rotate (under the ideal assumption that the hinge joint is modeled as a massless, frictionless rotational axis). Subsequently, one end of the wound electrode is fixed to the free end of the spring, and the electrode is wound around a prefabricated helical groove on the cylindrical core (pre-lined with electrodes and dielectric film). The other end of the wound electrode is anchored to segment 2, thus constraining its rotation. By applying different voltages, the friction between the two electrodes in the HWS-ELJ can be adjusted: higher voltages increase the friction and thereby increase the pulling force exerted on segment 2, requiring greater external force to bend the finger and thus enhancing stiffness; conversely, reducing the voltage decreases the stiffness, thereby achieving variable stiffness control.

\begin{figure}[h]
    \centering
    \begin{subfigure}{0.8\linewidth}
        \centering
        \includegraphics[width=\linewidth]{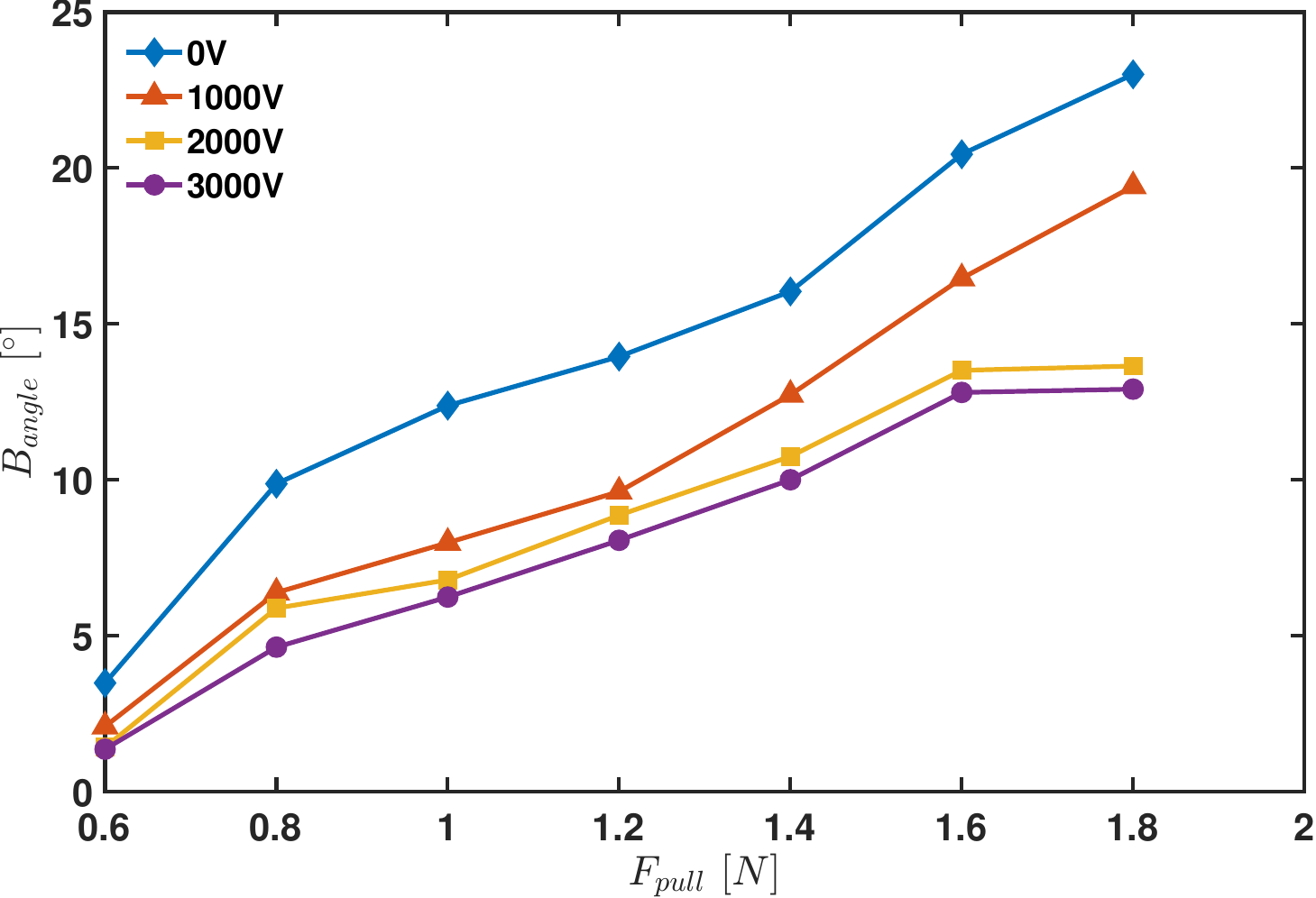}
        \vspace{-5mm}
        \caption{Diagram of Bending Angle vs. Voltage}
        \vspace{2mm}
        \label{fig:angle2}
    \end{subfigure}
    \begin{subfigure}{0.8\linewidth}
        \centering
        \includegraphics[width=\linewidth]{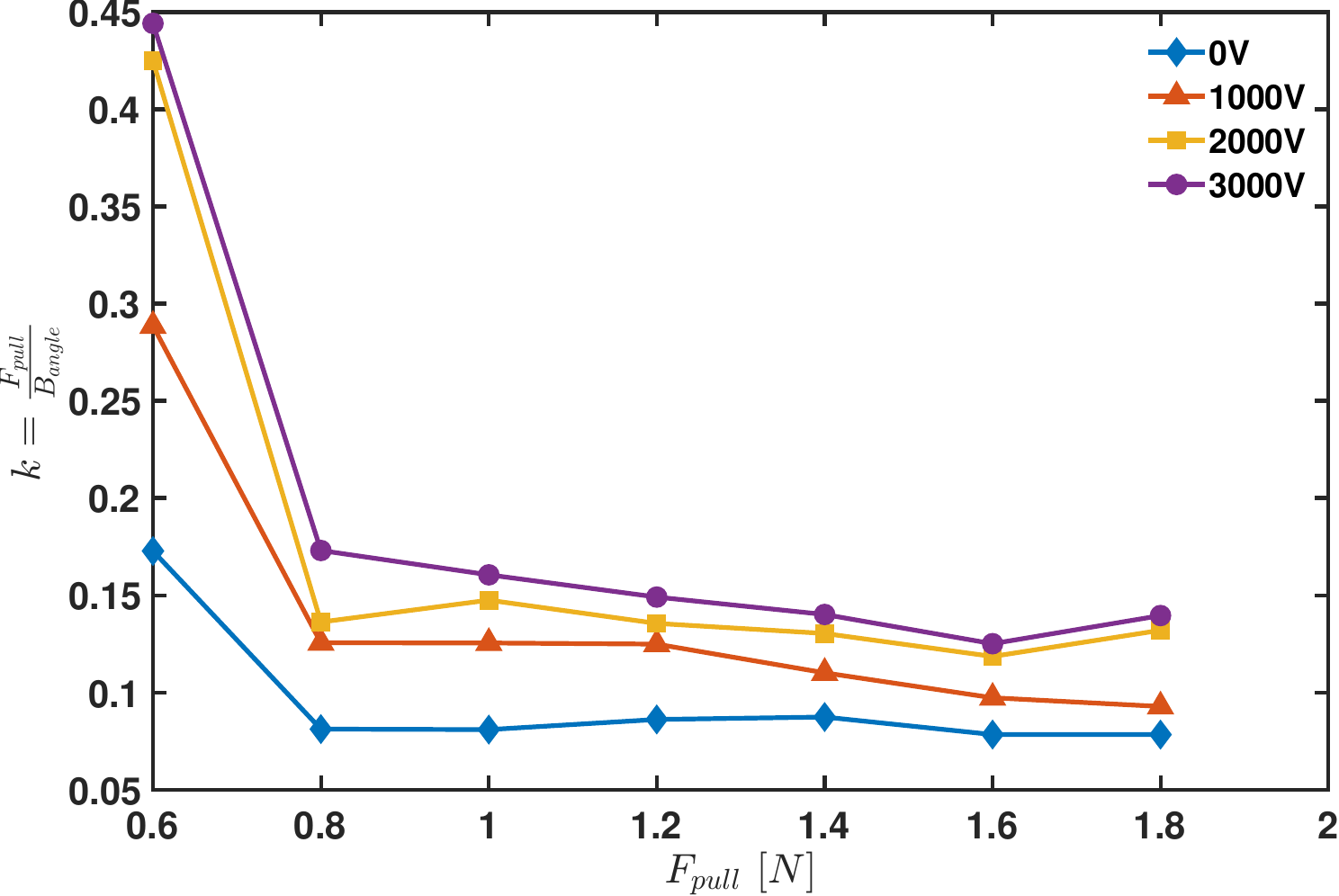}
        \vspace{-5mm}
        \caption{Diagram of the Defined Stiffening Factor vs. Voltage}
        \vspace{1mm}
        \label{fig:angle1}
    \end{subfigure}
    \caption{Diagram of Bending Stiffness vs. Voltage}
    \label{fig:angle}
    \vspace{-4mm}
\end{figure}

The spring preload is taken as an example to illustrate the stiffness modulation in practical applications. Under the ideal assumption, the torque at the joint is provided solely by the tensile force generated by the HWS-ELJ mechanism. In this experiment, electrodes with a width of 6 mm were helically wound with a pitch of 360 °. Under different applied voltages, various weights were loaded at the fingertip to evaluate the effect of HWS-ELJ on bending stiffness under different initial conditions. The experimental setup is shown in Fig.~\ref{fig:application}(b): the robotic finger equipped with the HWS-ELJ mechanism was fixed on the test platform, while the fingertip was connected by a thin string to a container for weights. To measure angular displacement of the fingertip under different loads and voltages, an electromagnetic(EM) sensor was mounted at the fingertip, with a generator positioned opposite to ensure accurate readings.

\subsection{Result Analysis}
When a mass of 180 g was applied, the deformation of the finger at 0 V and 3000 V is shown in Fig.~\ref{fig:application}(c). This result visually demonstrates that voltage regulation can effectively enhance the finger’s stiffness. Beyond this qualitative observation, Fig.~\ref{fig:angle}(a) illustrates the relationship between the net bending angle of the robotic finger sample and the externally applied vertical force under four voltage conditions. Here, the net bending angle is defined as the current bending angle minus the initial angle measured without external vertical loading. The experimental results indicate that the bending displacement increases with the applied load, while the entire curve shifts downward as the voltage increases, suggesting that the effective stiffness of the robotic finger specimen is enhanced with higher applied voltages.
To quantitatively evaluate the variation in bending stiffness induced by the HWS-ELJ mechanism, the effective stiffness of the robotic finger joint is defined as follows:

\begin{equation} 
k=\frac{F_{pull}}{B_{angle}}
\end{equation}

$F_{pull}$ denotes the vertical tensile load applied at the fingertip, and $B_{angle}$ denotes the corresponding bending angle at the joint under this load. Fig.~\ref{fig:angle}(b) presents the variation of the stiffness coefficient during bending deformation under different voltages. It can be observed that the stiffness coefficient increases significantly with voltage, further demonstrating the effectiveness of the HWS-ELJ technique for stiffness modulation.
Moreover, for the robotic finger specimen in this experiment, the decrease in $k$ within the larger loading range is more gradual compared to that in the smaller loading range. This phenomenon is primarily attributed to the initial force provided by the spring in the experimental setup. Under larger loads, the associated angular displacement is greater, which increases the spring extension and consequently the initial force it provides, thereby enhancing the intrinsic stiffness of the sample under large deformations. Consequently, the experimental observations are consistent with the theoretical expectations.

\section{Discussion and Conclusion}

This study introduced a novel variable stiffness mechanism based on a helically wound structured electrostatic layer jamming (HWS-ELJ), and systematically investigated its stiffness regulation mechanism and feasibility through theoretical modeling and experimental validation. The results demonstrate that the HWS-ELJ mechanism can achieve rapid and substantial stiffness variation by adjusting the applied voltage while maintaining a compact structure. Compared with conventional ELJ strategies, the proposed HWS-ELJ exhibits clear advantages, including reduced volume occupation, high modulation efficiency, and significant stiffness enhancement within a limited footprint. These features highlight its strong potential for applications in special robotic joints, robotic fingers, and wearable assistive devices where compactness and compliance adaptability are critical.

Meanwhile, limitations should also be noted. Localized low-pressure regions may occur at the contact interface, potentially leading to stiffness fluctuations.
In addition, since the approach relies on electrostatic adhesion, achieving substantial stiffness enhancement necessitates high-voltage operation, which makes careful insulation design essential; future improvements may be achieved through material-level optimization. Overall, despite these constraints, the proposed HWS-ELJ demonstrates promising potential for variable stiffness applications in robotics, which warrants further refinement and exploration in future studies.

\bibliographystyle{IEEEtran}
\normalem
\bibliography{IEEEexample}
\end{document}